\documentclass[
]{ceurart}

\usepackage{listings}
\usepackage{comment}
\usepackage{todonotes}
\usepackage{listings}
\usepackage{graphicx}
\graphicspath{ {./figures/} }
\usepackage{amsthm}
\usepackage{wrapfig}

\newtheorem{definition}{Definition}

\newcolumntype{d}[1]{D{.}{.}{#1}}
\begin{document}

\copyrightyear{2024}
\copyrightclause{Copyright for this paper by its authors.
  Use permitted under Creative Commons License Attribution 4.0
  International (CC BY 4.0).}

\conference{14th International Conference on Formal Ontology in Information Systems (FOIS 2024),
  08-09 July 2024 (online) and 15-19 July 2024, Enschede, Netherlands}

\title{The Mercurial Top-Level Ontology of Large Language Models}


\author[1]{Nele Köhler}[orcid=0009-0004-0532-0721, email=nele.koehler@ovgu.de]
\address[1]{Institute of Intelligent Cooperating Systems, Otto von Guericke University Magdeburg,
Germany}

\author[1]{Fabian Neuhaus}[%
orcid=0000-0002-1058-3102, 
email=fneuhaus@ovgu.de]
\fnmark[1]
\fntext[1]{Corresponding author: fneuhaus@ovgu.de}

\begin{abstract}
In our work, we systematize and analyze implicit ontological commitments in the responses generated by large language models (LLMs), focusing on ChatGPT 3.5 as a case study. We investigate how LLMs, despite having no explicit ontology, exhibit implicit ontological categorizations that are reflected in the texts they generate. The paper proposes an approach to understanding the ontological commitments of LLMs by defining ontology as a theory that provides a systematic account of the ontological commitments of some text. 
We investigate the ontological assumptions of ChatGPT and present a systematized account, i.e., GPT's top-level ontology. This includes a taxonomy, which is available as an OWL file, as well as a discussion about ontological assumptions (e.g., about its mereology or presentism). We show that in some aspects  GPT's top-level ontology is quite similar to existing top-level ontologies. However,  there are significant challenges arising from the flexible nature of LLM-generated texts, including ontological overload, ambiguity, and inconsistency.
\end{abstract}

\begin{keywords}
  ChatGPT \sep
  top-level ontology \sep
  large language model \sep 
  LLM 
\end{keywords}

\maketitle

\section{Introduction}

\begin{figure}
    \centering
    \includegraphics[scale=0.5]{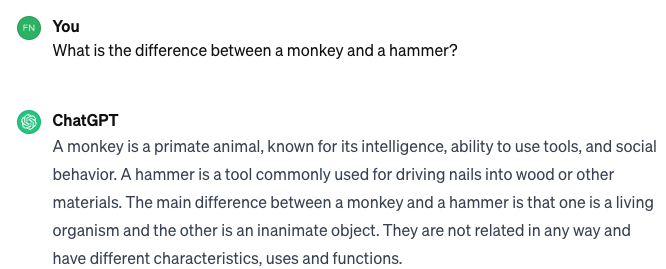}
    \caption{ChatGTP uses categories like `living organism' and `inanimate object'.}
    \label{fig:monkey}
\end{figure}

Large language models (LLMs) internalize knowledge about the world, which is reflected in the texts they generate. E.g., questions like \textit{What is the largest city in Western Europe?}  ChatGPT\footnote{Note, when we speak of ChatGPT in the course of the paper, we mean the interface that allows the use of the ChatGPT 3.5 model from OpenAI.} can answer reliably and correctly. 
While this knowledge is not represented explicitly in the form of an ontology or some other knowledge representation structure, the responses of LLMs involve implicit categorizations of entities in categories. E.g., the question \textit{What is the difference between a monkey and a hammer?} may elicit a response that contrasts two categories, namely a living organism and an inanimate object (see Fig.~\ref{fig:monkey}). Importantly, this distinction is introduced without any priming by the user like questions about ontological categories. 
Thus, while LLMs do not contain an explicit ontology, their responses contain implicit ontological commitments, which reflect the implicit ontological assumptions of the text corpora that the LLMs are trained on. 

In this paper, we will analyze these ontological commitments and present a top-level ontology that represents ontological distinctions that are made by ChatGPT 3.5 (Section~\ref{sec:topLevel}). As we will discuss in Section~\ref{sec:methodology} in more detail, because of the inherent differences between the technologies, the step from LLM to ontology is methodologically quite problematic.  
As we will discuss in Section~\ref{sec:discussion}, this common sense ontology differs from top-level ontologies from the literature (like BFO, DOLCE, UFO, etc.) and is certainly not as well developed, but there are recognizable similarities. However, there are significant differences. While terms in an ontology are associated with fixed compositional, model-theoretic semantics, LLMs are trained to produce tokens stochastically depending on context. Thus, while ontologies resolve ambiguities, LLMs reproduce them.  The mercurial nature of the terms used by LLMs is a significant obstacle to the investigation and the use of LLM's top-level ontologies.

Nevertheless, we believe they are interesting to study for two reasons: since the ontological distinctions made by LLMs are a distillation of the ontological distinctions made by the authors of the millions of texts that the LLMs are trained on, this top-level ontology may be considered as an approximation of the common sense ontology that underpins everyday discourse. To study this common sense ontology is of interest in itself. Further, there are already various efforts to use LLMs in the ontology development process \citep{ciatto2024large}, and these tools will likely become a staple in ontology engineering. Understanding the ontological assumptions of LLMs will make it easier to integrate the output they produce within the context of top-level ontologies like BFO, DOLCE, or UFO.

\section{Related Work}
\label{sec:related work}

We aim to develop the underlying top-level ontology of ChatGPT, that covers the most important and basic categories.
There are already numerous manually created top-level ontologies of high quality. \citep{Partridge_2020} present and compare 37 different TLOs in their overview. Some well-known TLOs are BFO, DOLCE, GFO, UFO, and GMO. These top-level ontologies differ methodologically and philosophically. BFO by \citep{arp_building_2015, smith_2015}, for example, embraces ontological realism, while DOLCE by \citep{gangemi2002sweetening} is more focused on conceptual and linguistics aspects of ontology engineering, and GMO is only linguistically motivated  \citep{bateman1995generalized}. Other approaches, such as UFO, attempt to synthesize the ideas of the TLOs \citep{guizzardi2015towards}. UFO emerged from a synthesis of DOLCE and GFO, a multicategorial approach by \citep{herre2010general}. TLOs can ultimately form a basis for specific domain ontologies, which is why SUMO, for example, is very broadly diversified to offer as many points of contact as possible \citep{niles2001towards}. 
While the different ontologies represent different 
These TLOs are characterized by the fact that they were developed manually, are of high quality, and are philosophically sound. 

Our work differs from the existing work on TLOs since our goal is not to manually create a new TLO, but to investigate the top-level ontology of ChatGPT. ChatGPT is based on the GPT-3.5 model, which is a kind of LLM or, more precisely, a generative pre-trained transformer \citep{Openai}. Other LLMs are for example Microsoft Copilot \citep{MicrosoftCopilot}, Google Bard \citep{GoogleBard}, and Claude \citep{IntroducingClaude}. 

LLMs are already used for ontology development and there are already many different high-quality and well-founded research approaches, for example using it for combining terms for representing domain entities \citep{lopes2023using}, enriching ontologies with a fine-tuned GPT-3 model as a tool \citep{mateiu2023ontology} or exploring the facilitation of (semi-)automatic construction of knowledge graphs, through open-source LLMs\citep{kommineni2024human}. But there are many other 
projects and approaches to support ontology development using LLMs \citep{pan2023unifying, caufield2023structured, LLMs4OL, chen2023large, zhao2024using, 10.1145/3587259.3627571}. That the use of LLMs for ontology development is  of high research interest is also supported by the fact that there are currently several venues dedicated to the topic.\footnote{There is a special issue of Applied Ontology on "Ontologies and Large Language Models"   \url{https://www.iospress.com/catalog/journals/applied-ontology}, a special track  at ESWC \url{https://2024.eswc-conferences.org/call-for-papers-llms/}, and a challenge at ISWC \url{https://www.nfdi4datascience.de/news/202402_llms4ol2024/}}
The existing research is about using LLMs to support the development of domain ontologies.
In contrast, we focus on the top-level ontology of ChatGPT. 

\section{Methodology }
\label{sec:methodology}

The responses of LLMs may contain ontological commitments. 
For example, the response of ChatGPT in  Fig.~\ref{fig:monkey} uses ontological categories of living organisms, inanimate objects, characteristics, abilities, and (social) behavior. Further, while it is not stated explicitly, the phrasing seems to indicate that living organisms and inanimate objects are disjoint, and ChatGPT asserts a `is used for' relationship between a kind of inanimate object and its function. 
In Section~\ref{sec:topLevel} we will systematize these assumptions and present the result as `GPT's top-level ontology'.
 However, this goal raises several important methodological concerns. Most importantly, is there such an ontology in any meaningful sense of the word?

To address this question we need to be aware that the term `ontology' is used ambiguously in the literature:  
An ontology$_D$ is a kind of document, (e.g., version 228 of the ChEBI ontology$_D$ that was released in January 2024 and can be downloaded on Bioportal).  An ontology$_C$ is a conceptualization of a certain domain by a person or a group of people (e.g., the ontology$_C$ of human anatomy according to Galen). (The distinction  between these two senses of `ontology' is made, for example, in \citep{guarino1998formal}.) 
Further, an ontology$_R$  consists of the categories of entities and their relationships to a certain part of reality  (e.g., the ontology$_R$ of chemistry) \citep{smith2004}.%
\footnote{ 
The three different notions of ontology are connected. Typically, an ontology$_D$ is a representation of some ontology$_C$ but is intended to be an accurate representation of ontology$_R$. In the case of Galen's ontology$_C$ and the ontology$_R$ of human anatomy, the gap between conceptualization and reality is quite large. As science progresses this gap, hopefully, decreases. }

Strictly speaking, a LLM uses an ontology in neither of these senses. LLMs have no access to reality beyond the documents that are in their training corpora and, thus, have no access to ontology$_R$. Since these corpora may include ontologies$_D$  among the millions of other documents,  ontologies$_D$ may be used to train LLMs. However, these ontologies$_D$ are not treated any differently to other documents in these corpora, e.g.,  Magna Carta.  After training is complete the LLM does neither contain a copy of the Magna Carta nor any ontology$_D$. 
  
Further, LLMs are like stochastic parrots, which produce without any comprehension texts that are ``not grounded in communicative intent, any model of the world, or any model of the reader’s state of mind'' \citep{bender2021dangers}. A parrot may be trained to say `There are possible worlds!', but we would not consider it a modal realist, since it lacks both the communicative intent as well as the understanding of the concepts involved. The same is true for LLMs, and, thus, it would be a category mistake to attest LLMs to a conceptualization or an ontology$_C$. 

One could argue that while the parrot lacks understanding of its utterances, they reflect the  ontology$_C$ of its trainer. Analogously, one might argue, the LLM is trained by processing millions of documents and, thus, learns an ontology$_C$ that coagulates the ontological views of the millions of authors of these documents. However, this is unfortunately not the case. LLMs generate text one token at a time based on a function that maps sequences of tokens to probability distributions of tokens. Consequently, the sentences generated by a given LLM are dependent on their prompts and even minor changes to the prompt may lead to the generation of contradictory sentences.  Hence, the texts generated by a LLM are often not even logically consistent, let alone express some coherent ontological view \citep{neuhaus2023ontologies}.  Thus, LLMs do not contain an implicit ontology$_C$  within their billions of parameters, which we can induce the LLM to reveal by providing the appropriate prompts.

In the remainder of this paper, we use `ontology' in a fourth sense, namely as a theory of ontological commitments. More precisely: 
\begin{definition}\label{def:ontology}
Let $T$ be a text (i.e., a sequence of sentences) and $K$ a category of entities. $T$ is \emph{ontologically committed} to $K$s if and only if $T$ logically entails that $K$s exist. An \emph{ontology} of \ $T$ is a set of sentences about the relationships between the categories that $T$ is ontologically committed to, which is -- as far as possible -- logically consistent with $T$. 
\end{definition}  
Definition~\ref{def:ontology} is based on a variant of an entailment account of ontological commitment \citep{sep-ontological-commitment}. 
An ontology is a theory (i.e.,  set of sentences) that provides a systematic account of the ontological commitments of some text $T$.\footnote{In applied ontology, these sentences are usually sentences in a formal language (e.g., OWL DL) together with annotations that specify the intended semantics of the vocabulary with the help of annotations.} 
Since a text is defined as a sequence of sentences, `text' does not imply the written form, but includes spoken statements by (possibly different) domain experts.  An ontology usually includes a taxonomic hierarchy and disjointness relationships between the categories. However, it may contain more complex assertions about the relationships between the categories, for example, a statement of the form: all instances of $K$ that meet some condition $C$ stand in relation $R$ to some instance of $K^\prime$. Thus, the vocabulary of the ontology is not restricted to unary predicates that denote the ontological commitments of $T$ but may include additional predicates, e.g., to express relationships between instances of categories.  $T$ may reflect different points of view and, thus, may be logically inconsistent. For this reason, we do not assume that an ontology of $T$ is logically consistent with $T$, but it should minimize the number of sentences in $T$ that it is logically inconsistent with. 

Definition~\ref{def:ontology}  has the benefit of being applicable to the texts generated by LLMs without having to assume that LLMs can form concepts although LLMs produce inconsistent texts.
The goal of this paper is to provide a systematic account of important ontological commitments in the text generated by ChatGPT. 

This task is complicated by the fact that the user may influence the ontological commitments in generated texts by the prompt that is used. For example, if one asks ChatGPT to explain the difference between particularized properties and tropes, it will generate a text that contains ontological commitments to particularized properties and tropes. Analogously, by providing the appropriate prompts one can induce ChatGPT to provide a Platonian or an Aristotelian account of the nature of change. 
Since LLMs are trained on text corpora that contain philosophical texts,  a user may prompt ChatGPT to generate texts reflecting a vast array of philosophical perspectives, which contain equally diverse ontological commitments. 
However, for the purpose of our paper, we are not interested in the ontological commitments of texts that reflect philosophical positions in the literature. 
Since our goal is to study ChatGPT's ontological commitments and assumptions, which may influence its usefulness as a tool for ontology engineering, we use prompts that are designed to reveal ontological commitments without priming ChatGPT to respond based on the philosophical literature. Another difficulty is that, in contrast to a human, ChatGPT has no introspection about its ontological commitments. 
E.g., a prompt like ``What are the most general ontological categories that you use to organize your knowledge?'' will prompt it to generate an answer about ontological categories from the literature, but according to our observation, these are not necessarily the ones it actually uses.\footnote{Note that minor variants of this prompt will lead to different responses, thus, supporting the claim that ChatGPT's response does not reveal the categories it actually uses. E.g., compare 
\url{https://chat.openai.com/share/655415ca-0479-45ba-937e-331527dd3c61} with 
\url{https://chat.openai.com/share/cd65d085-7d5f-422a-bda2-51260c8799d8}.
}

Therefore, to study ChatGPT's ontological commitments, we use a more indirect approach by asking questions like in  Fig.~\ref{fig:monkey}, which illicit responses that contain ontological categories (e.g., animate object and inanimate object). These categories we use in follow-up questions about similarities of entities in these new categories (e.g., ``What is the difference between the monkey and monkey's behavior?'') and other possible instances of these categories (e.g., ``Are shadows inanimate objects?''). We also use more theoretical questions about the relations of the categories (e.g., ``Are there objects that are both animate and inanimate?''), however, these sometimes lead to responses by ChatGPT that are incoherent with follow-up questions. E.g., in the same session, when asked whether there are entities that are both physical and abstract, ChatGPT claimed that national flags exhibit both physical and abstract characteristics. However, when asked whether a national flag is a physical object or an abstract object, it responds that it is physical and not abstract. This example illustrates the points made above: the texts that are generated by ChatGPT are neither logically consistent nor informed by introspection. For the same reason, asking ChatGPT for definitions is not as helpful as one might hope. E.g., it defines ``physical entity'' as ``entities that have a tangible, material existence'' and ``abstract entity'' as ``entities that lack a tangible, material existence''. Thus, according to these definitions, the categories should provide a partition of entities. However, as just mentioned, it sometimes claims that some entities are both abstract and physical. Further, it also claims (at least sometimes\footnote{There is some significant variability depending on the exact formulation of the prompts.})  that shadows lack tangible, material existence but are not abstract entities.

These examples illustrate that to understand the ontology of ChatGPT, asking it for definitions of the categories it uses, frequently leads to misleading responses. For this reason, we study ChatGPT's ontology primarily by prompting it to use categories for the classification of examples or distinguishing categories. 

A particular challenge for studying the ontology of LLMs is that only minor changes to prompts may lead to different results.\footnote{
In the case of ChatGPT even asking exactly the same question may lead to different results, since (a) ChatGPT is configured to respond non-deterministically (because its `temperature' is larger than zero) and (b)  previous interactions within a given session may influence the response to the next input by the user.
} 
To ensure that the ontology we present in the next section reflects the typical ontological commitments of texts produced by ChatGPT, we studied the results of asking similar questions in different variations. 
We only included categories that were used consistently in different contexts. However, since ChatGPT generates text based on a stochastic process, it may produce texts that are inconsistent with the ontology we present in the next section. This is obviously a challenge for the verifiability of our claim that the  ontology presented in Section~{sec:topLevel} reflects the ontology of ChatGPT. Thus, to ensure at least transparency, we published the transcripts of our interactions with ChatGPT, which our claims are based on.%
\footnote{Our main chats can be accessed here: 
\url{https://chat.openai.com/share/23d34cc5-04b4-4dd6-aaae-99994d1ca4c9}, \url{https://chat.openai.com/share/af5f19b7-8f00-4567-adc5-9b8977124048}, and 
\url{https://chat.openai.com/share/43037f0a-3626-4ed7-ad23-d056609819b7}.
Note, that in the course of the paper, we sometimes also use smaller chats for isolated examples. We also provide the link to the chat history at the appropriate passages.}

\section{GPT's top-level ontology}
\label{sec:topLevel}

\subsection{Hierarchy}
\begin{wrapfigure}{l}{0.35\textwidth}
    \centering
    \includegraphics[scale=0.5]{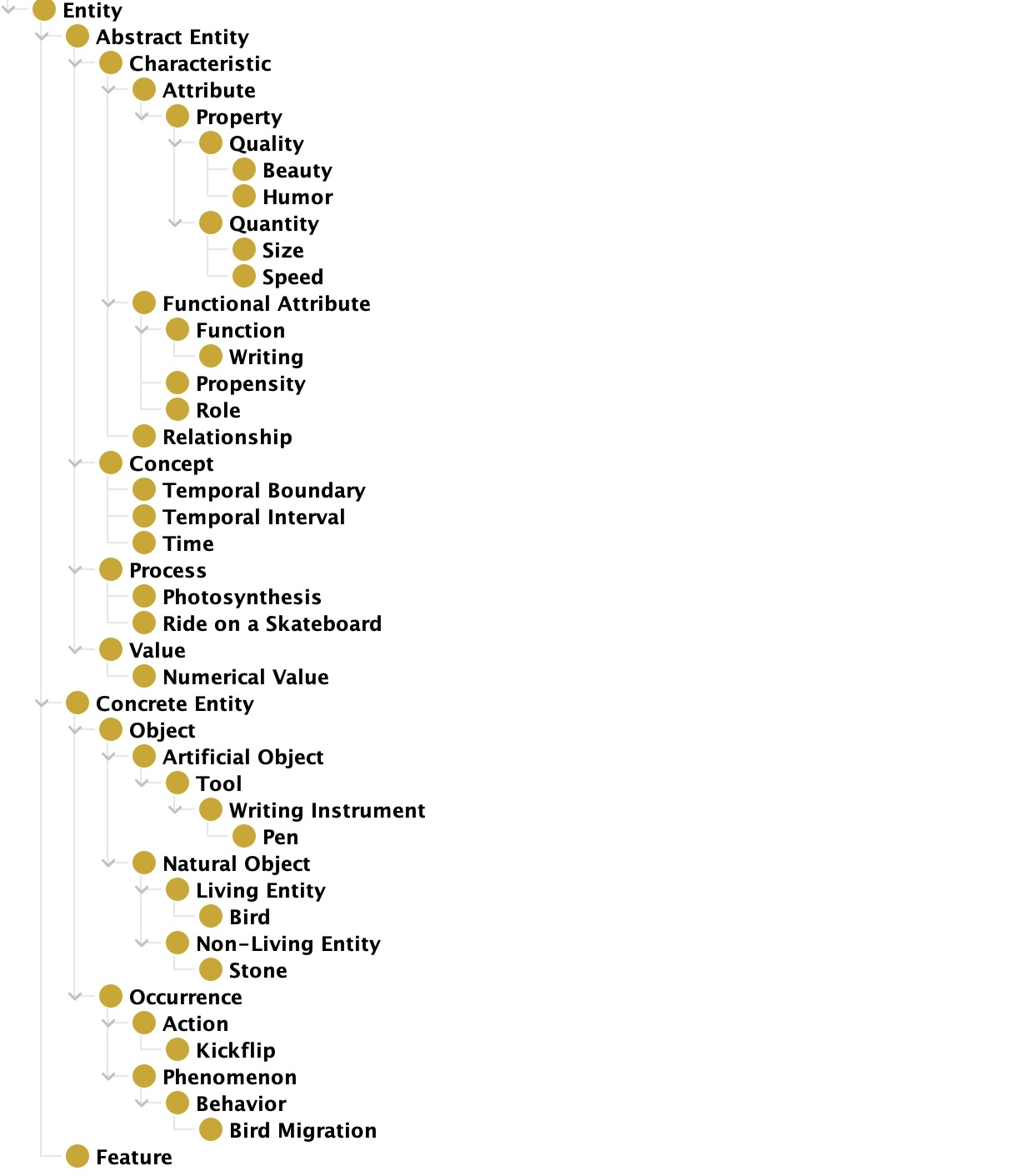}
    \caption{Hierarchy of ChatGPT} 
    \label{fig:hierarchy}
\end{wrapfigure}

In this part, we present the ontological categories, which we have isolated in the conversations with ChatGPT. As already discussed in Section~\ref{sec:methodology}, ChatGPT does not use fixed definitions of its terms and it sometimes provides what seems to be conflicting or outright contradictory responses. However, there is a stable core of ontological commitments that are consistently made by ChatGPT (with few exceptions). This is the source, which we used to construct the ontology that we present in this section.  Fig.~\ref{fig:hierarchy} shows the subsumption hierarchy of the top-level categories and some additional classes, which illustrate the meaning of the higher-level categories.  The ontology is available at \url{https://w3id.org/gptto/v1.0.0/>}. 
We will discuss the limits of this approach in Section~\ref{sec:discussion}.

The most general category ChatGPT uses is \textit{entity}. E.g., in response to questions like \textit{What do a pen and a bird have in common?}, they are compared as entities with different characteristics.
Furthermore, it differentiates between \textit{abstract} and \textit{concrete entities}. The terms for this can vary in different conversations, but the subdivision into the concepts of the categories remains roughly the same. Concrete entities, which are also called \textit{physical entities}, are characterized by the fact that they physically exist, and have perceivable, observable, or measurable characteristics or behaviors. These entities can be part of complex systems or networks and interact with one another and with their environment through processes.  

In contrast, \textit{abstract entities} are those that do not exist in a physical form and have no physical attributes, therefore they cannot be perceived directly through the senses. These include concepts, values, processes, and characteristics. Abstract entities are disjoint from concrete entities.
In addition to abstract entities and concrete entities, there are also \textit{features}. These are entities that are used to describe or identify instances in more detail. Features can be abstract or concrete.

The category concrete entity is divided into \textit{objects} and \textit{occurrences}. An object is a concrete entity that exists physically and has physical characteristics, that can be perceived. Objects can be \textit{natural entities} and \textit{artificial objects}. There is a deeper distinction between artificial objects, that are made by human beings and do not naturally occur in the environment but are created for a specific purpose or function. These include, for example, tools. Natural objects, on the other hand, are concrete entities of the natural world that exist without direct human intervention or manufacturing. Instead, they arise through natural processes. The natural objects are divided into \textit{living entities} (e.g., birds) and \textit{non-living entities} (e.g., stones).

\textit{Occurrences} are concrete entities that happen in the physical world and can be perceived, observed, or measured. An example would be the flying of a bird past my window or a kickflip, which can be classified as an \textit{action}. Actions occur as sets of movements of living entities.

Another subclass of occurrence is \textit{phenomenon}.\footnote{
ChatGPT uses the term `phenomenon' ambiguously, but most of the time it is used to denote a kind of occurrence; e.g., 
\url{https://chat.openai.com/share/77a200c5-6a4b-4304-a152-8fb275acbdbc}
}
Therefore, phenomena are occurrences that imply some kind of pattern or broader significance, which then can be scientifically studied. The behavioral pattern of birds flying south during winter can be classified even more precisely, namely as \textit{behavior}. Behaviors are phenomena of living entities.

While concrete entities primarily comprise objects and occurrences, the class of abstract entities is subdivided into concepts, values, processes, and characteristics. 
\textit{Concepts} are abstract entities, that exist as ideas or mental representations and do not have a physical presence. These include, for example, time or the concept of a bird. Concepts can influence human life and society and can represent concrete entities, but they do not themselves exist as physical objects. 

\textit{Values} are abstract entities that are used to express certain attributes or concepts. They can be based on qualitative or quantitative aspects. For example, a quality like beauty can be expressed as an aesthetic value or moral values that express a certain understanding of moral. \textit{Numerical values} are connected with properties, since they represent the magnitude of quantities. 

\textit{Processes} as a subclass of abstract entities encompass sequences of actions, changes, or transformations that occur over time, e.g. photosynthesis or the ride on a skateboard. Processes are manifested through individual actions. Below in section ~\ref{sec:discussion} we will discuss the relationship between processes and actions further, especially concerning the fact that processes are abstract entities and actions are concrete.  

\textit{Characteristics} are abstract entities that help to identify, classify, or describe classes of entities or individuals in terms of typical behavior. What distinguishes characteristics from features is the fact that features describe instances in more detail, whereas characteristics describe aspects of classes or typical behavior of individuals. 'My apple is round' is, therefore, an example of a feature of my apple, i.e. an instance. 'Apples are round', on the other hand, would be a characteristic of apples in general, i.e. the class of apples. The roundness can therefore be both a feature and a characteristic. Jack's dark humor, for example, does not concern a class, but Jack as an individual. Nevertheless, dark humor is a characteristic here because it describes the phenomenon of typical behavior that occurs in Jack.

\textit{Attributes} are characteristics that can be observed, measured, or represented. They seem to contribute to observing more overarching matters. For example, someone's dark humor is a characteristic, that describes a part of their personality. Furthermore, the observation of humor can also be representative of behavior or communication style, in which case it is considered an attribute.

\textit{Properties} are objective, measurable attributes or characteristics of an object or entity that can be quantified, observed, or described using empirical methods. Properties can often be represented in the form of data. In this respect, humor would be difficult to consider as a property, but the speed of an eagle would be something that can be measured and observed and can be collected as data.
They can be divided into \textit{qualities} e.g. colour or beauty and \textit{quantitative attributes} e.g. the specific saturation of a colour, speed, or size. Qualities are subjective, perceptual, or experiential attributes of an entity that are subjectively perceived or experienced by individuals, while quantitative attributes comprise numerical or quantitative measurements of the properties of an object. Both can be represented by corresponding values, whereby values for quantities are created by measurements, and values for qualities are more abstract and conceptual. The classification of qualities and quantities in the hierarchy is not easy because it always depends on the context and whether they are used to identify an entity, are observed, or can be recorded as data. 

\textit{Relationships} are characteristics that refer to the relationships or connections that an object has with other objects or entities. 

\textit{Functional attributes} include functions, roles, propensities/dispositions, or purposes that an object fulfills within a system or context. Examples include the function of a tool (e.g. cutting, measuring, writing), and the role of a component in a machine. \textit{Propensities} are functional attributes that refer to the tendency of a concrete entity to behave or act in a specific way.

Apart from the classes, we have also tried to recognize \textit{relations} by isolating frequently occurring terms. This was much more difficult because even if terms occur frequently, it is not always possible to identify a clear relation or it is difficult to determine which entities are part of the relation. 
One relation that we did notice, however, is \textit{manifest}. It is often used for the relationship between abstract and concrete entities. For example, the action (concrete entity) manifests the process (abstract entity)


\subsection{Further ontological assumptions}
In \textit{A survey of Top-Level Ontologies - To inform the ontological choices for a Foundation Data Model}, the authors have summarized the results of a broad survey on ontologies conducted by the IMF technical team. It discusses the important ontological choices that are relevant for top-level ontologies \citep{Partridge_2020}. Not all of the choices listed apply to the ChatGPT-based ontology, but many do. 
Thus, we discuss ontological assumptions contained in ChatGPT's ontology based on the criteria in \cite{Partridge_2020}. 

\textbf{What structural relations are used and what are their properties?} 
\cite{Partridge_2020} mention three  structural relationships:
type-instance, subtype-supertype, and part-whole.

ChatGPT uses all of these relations. 
As Fig.~\ref{fig:monkey} illustrates, ChatGPT uses subtype relationships in its answers (e.g., a hammer is a kind of tool).  
The subtype-supertype relationship is distinguished by ChatGPT from 
\textit{type-instance} relations.\footnote{For example, ChatGPT treats the former as transitive, the latter as intransitive. 
\url{https://chat.openai.com/share/dd7a7742-b232-4895-b690-da2b72c5a793}
}   
ChatGPT also uses \textit{part-whole relations}, for example, processes are composed of individual actions. However, part-whole relations occur less frequently in ChatGPT's responses than in the other two relationships.  
Other ontological interesting relations used by ChatGPT are `consist of'  and `involve', and `manifestation of'. The latter relation holds between concrete entities and abstract entities. 

Since ChatGPT does not use a formal ontology that includes axioms, and its use of terminology is fluid, it is difficult to identify the formal properties of these relationships as discussed in \citep{Partridge_2020}. One criterion that ChatGPT meets is that the subsumption hierarchy is upward bounded in the sense that `entity' is the top category that subsumes all other categories. 

\textbf{Does it use 
General Extensional Mereology (GEM)?}
GEM is a strong mereological theory. One of its features is that there is a (unique) sum of any two arbitrary entities. 
According to ChatGPT, no object is the mereological sum for any given two objects. E.g., if a duck sits on a table, no object is the sum of both. 
Thus, concerning objects ChatGPT does not embrace GEM.  
However, when asked directly, it suggests that there may be a conceptual construct `duck/table'. In contrast to ducks and tables, this construct is an abstract entity in ChatGPT's ontology, thus, it is no tangible object in the physical world.
This example illustrates that for ChatGPT concrete entities may be part of abstract entities and that different mereologies may applied to different categories of entities. 

\textbf{If the location of two objects overlap, do they share parts?} 
  \cite{Partridge_2020} define \textit{interpenetration} as follows: two objects interpenetrate when they do not share parts, but their locations overlap. E.g., if we assume that the statue and the clay it is made of are two different entities and that the statue consists of the clay, but has no clay as part, then this would be a case interpenetration. A contrasting view would be that the statue and the lump of clay share parts, and, thus, there is no interpenetration.   
  In a conversation with ChatGPT a flower bouquet made out of a LEGO set is not seen as one object, but as two objects whose parts are interconnected. If a part of the LEGO set falls off, only the set is incomplete, the bouquet is still intact, even if the arrangement has changed somewhat due to the missing LEGO brick. Thus, ChatGPT's ontology seems not to embrace interpenetration.\footnote{The examples can be found in the following chat histories:
\url{https://chat.openai.com/share/63bf3995-0930-4b11-969b-f5f6819b1a21} 
\url{https://chat.openai.com/share/d5014140-1dad-40ab-baf7-541542c7e1b6} 
}

 \textbf{Does the ontology embrace endurantism or {perdurantism}?}
One important ontological distinction is between \textit{endurantism} and \textit{perdurantism}. According to endurantists, material objects are three-dimensional entities that persist through time and which are wholly present at any time they exist. According to perdurantists, material objects are four-dimensional entities that occupy space-time. Hence, at any given moment, only a temporal part of the material object is present. 
 While endurantists, typically, claim that there is a fundamental ontological difference between a material object and processes (e.g., John and John's life), perdurantists would, typically, deny that these are two different ontological categories. 
ChatGPT embraces endurantism because it distinguishes systematically and reliably between objects and occurrences as disjoint categories. There are some boundary cases, e.g.,  ChatGPT provided conflicting answers to the question of whether glaciers are objects or occurrents.\footnote{Use for comparison
\url{https://chat.openai.com/share/23d34cc5-04b4-4dd6-aaae-99994d1ca4c9} 
and
\url{https://chat.openai.com/share/af5f19b7-8f00-4567-adc5-9b8977124048} 
}  
But even in these cases  ChatGPT embraced only one option in a given session and did not claim that glaciers (and by extension other material entities) are both objects and occurrences at the same time. Further,   
 ChatGPT also distinguishes between a person and the person's life as two separate entities, where the former is an individual who experiences the latter.\footnote{\url{https://chat.openai.com/share/a77e6261-b629-456a-a3e8-dc4f92098899}}

\textbf{Is materialism adopted?}
TLOs may also divided by the question, of whether the world consists only of material things that exist in space and time or whether there are also abstract particulars, i.e., a distinction between \textit{materialism} and \textit{non-materialism} \citep{Partridge_2020}.  ChatGPT describes the world not only based on material entities but also immaterial entities. Therefore, ChatGPT embraces non-materialism.

\textbf{Are possibilia or possible worlds adopted?}  
ChatGPT does not talk about possibilities in other possible worlds in its answers, but everything is based on information from the actual world. When ChatGPT is asked about possible worlds, it treats them as hypothetical entities and addresses their existence in philosophical discourse. 
Therefore, there are no possibilia, possible worlds, or possible situations in ChatGPT's ontology. However, ChatGPT is committed to propensities and functions, which provide a kind of teleological vocabulary.

\textbf{Is the ontology presentist or eternalist?} 
Natural language is often tensed. For example, in `Yesterday, I was in Paris, today I am in Berlin, tomorrow I am in Warsaw', we use different variants of `to be' and the words `yesterday', `today', and `tomorrow' to distinguish past, present, and future. Presentism assumes that these distinctions have ontological significance, and that, strictly speaking, only present entities exist.  In contrast, eternalists claim that past, present, and future entities exist.\footnote{\cite{Partridge_2020} present eternalism and presentism as the only alternatives, but other positions on the reality of time are available. For example, according to 
`growing block theory'  the past and the present exist, but the future does not.}  

 If ChatGPT is asked about dinosaurs, it answers they existed in the past. Hence, ChatGPT uses tensed language to distinguish between past and present.   
 However, since ChatGPT is trained to generate well-written English sentences, it follows linguistic conventions. 
Thus, it is quite difficult to determine whether the tensed language is just a stylistic choice or an ontological commitment to presentism. 
 E.g., if asked whether (the late) Kirk Douglas is the father of Michael Douglas, ChatGPT answers affirmative. If asked, whether this entails that Kirk Douglas exists, it provides conflicting responses: first, it claims that this fact entails that Kirk Douglas exists, then it claims that he `no longer exists in the physical sense, but his legacy lives on through his work in film and the memories of those who knew him'.\footnote{\url{https://chat.openai.com/share/ffcaf7d2-3fa5-4481-9308-0950f57ac7ce}} It seems to us that ChatGPT's notion of `existence' is quite mercurial, and it switches back and forth between a presentist and an eternalist point of view. 

\textbf{Expressivity: Are indexicals `here' and `now' and are relations of arity larger than two supported?} Since ChatGPT can communicate in English, it supports indexicals and sentences with predicate phrases that express relations between more than two entities (e.g., `Berlin is between Paris and Warsaw.'). However, the ontology we created (based on  ChatGPT's output) is written in OWL DL and, thus, supports neither.  


\section{Discussion}
\label{sec:discussion}

In the previous chapter, we presented ChatGPT's top-level ontology, i.e., a hierarchy of categories (represented in an OWL file) and ontological assumptions. 
However, as pointed out in  Section~\ref{sec:methodology},  ChatGPT does not use an ontology (in the sense of a file), what we presented is an attempt by the authors to provide a systematic account of the ontological commitments 
that we identified in texts that were generated by ChatGPT. In this section, we discuss the similarities and differences to `normal' TLOs. 

In some aspects
ChatGPT's top-level ontology is surprisingly traditional: it contains a hierarchy of categories, which are instantiated by instances. The distinction between living and non-living entities has been established since  Aristotle's Categories \citep{barnes1984complete}. Many of the other categories are similar to the ones used in established ontologies. 
E.g., similarly to BFO, the ontology contains objects, that may participate in occurrences, that have properties (qualities in BFO), may play roles, and may have propensities (dispositions in BFO) or functions.
ChatGPT embraces dualism since it is ontologically committed to physical entities and mental entities, like mental constructs, mental actions, or mental processes.
By our comparison, we do not wish to suggest that ChatGPT's top-level ontology is similar to BFO. Quite contrary, there are major differences with respect to the categories, their definitions, and their organization in the subsumption hierarchy. E.g., ChatGPT's primary distinction is between abstract and concrete entities, while BFO does not include a category for abstract entities. We only want to point out that somebody familiar with established ontologies, is able to recognize familiar ontological distinctions by ChatGPT.

The major difference between ontologies and ChatGPT is that ontologies usually are carefully constructed in a way that they resolve ambiguities by distinguishing different concepts and providing clear definitions.  
 This is particularly important in the case of polysemes, where humans rely on context to disambiguate (if necessary). E.g., an ontology of cattle will introduce two different terms for `cow', one in the sense of bovine and the other for female bovine. Similarly, ontologists will distinguish between hammer-as-object, hammer-as-process, and hammer-as-function.    
 In contrast, LLMs generate texts based on a given prompt and a function that maps tokens to probability distributions of tokens. 
 Hence, LLMs do not disambiguate words but rather learn to use them appropriately in a given context. 
This mercurial use of language is a great benefit for the task of generating natural language texts,  but it is an obstacle for using LLMs for the task of creating ontologies. 

Firstly,  it leads to a kind of `ontological overload'
E.g., while an ontology typically would distinguish between a coin and the value it represents, ChatGPT treats the same entity as concrete and as abstract. In this example, ChatGPT states that fact explicitly\footnote{\url{https://chat.openai.com/share/5604931c-4319-4351-b485-98dd4f17c0b7}}, however, often it is more vague and claims that some entities have both physical and abstract characteristics.  E.g., the roundness of an apple is both a concrete feature of the apple and, at the same time a generic characteristic (an abstract entity). This phenomenon is not limited to the distinction between concrete and abstract entities, the entities in ChatGPT's ontology may be instances of different ontological categories that one expects to be disjoint. Depending on the context, they are treated to be one or the other.

\begin{figure}
    \centering
   \includegraphics[width=0.9\textwidth]{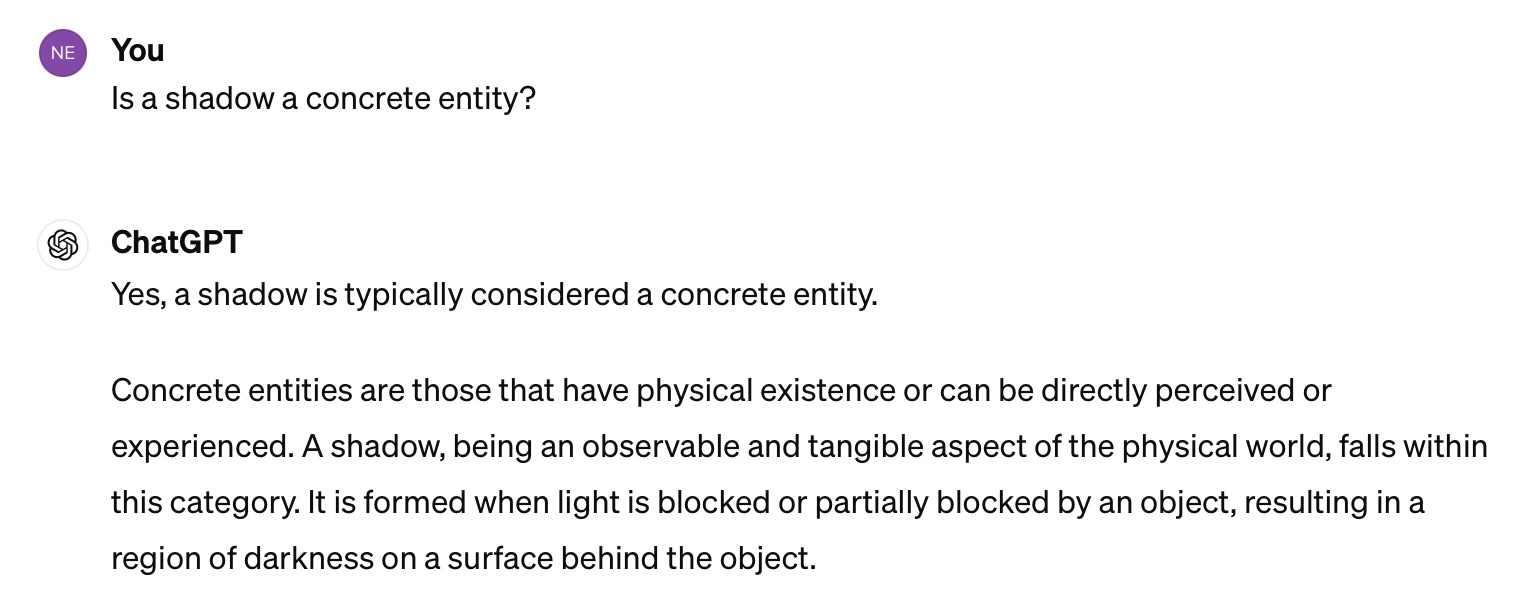}
    \caption{Classification of a shadow as a concrete entity}
   \label{fig:shadowconcreteentity}
\end{figure}
 
\begin{figure}
    \centering
    \includegraphics[width=0.9\textwidth]{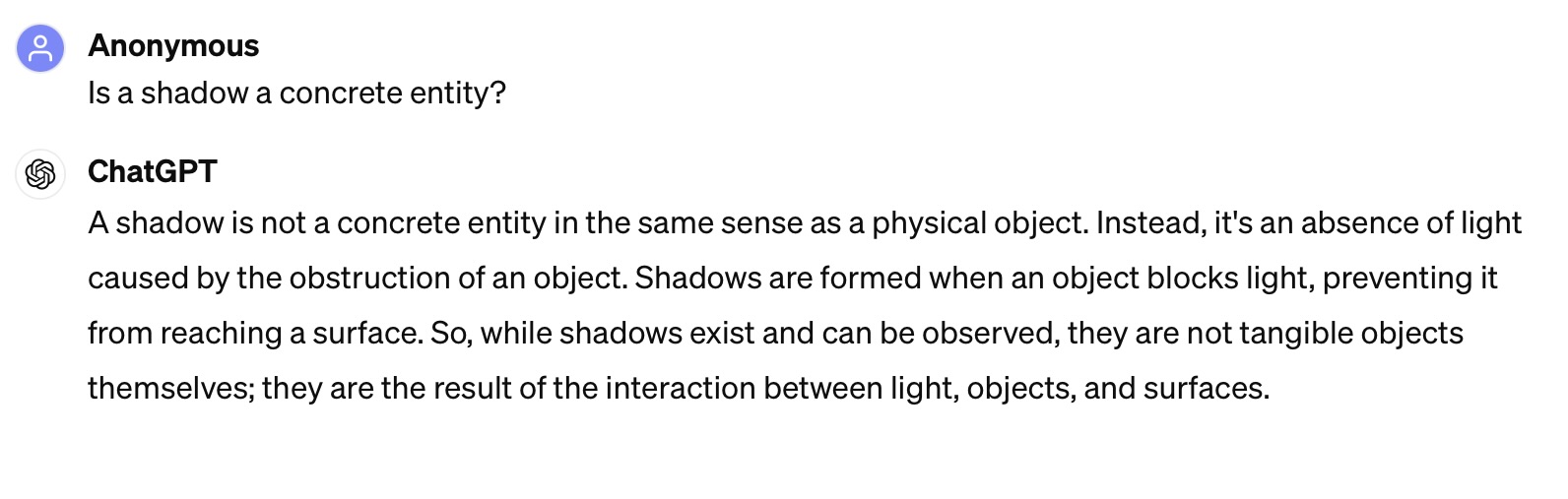}
    \caption{Classification of a shadow distinct from a concrete entity .}
    \label{fig:shadownotconcreteentity}
\end{figure}

 Secondly, it leads to inconsistent responses. 
 For example, in two separate conversations, ChatGPT may answer the same question in exactly the opposite way. For example, a shadow is classified as a concrete entity in  Fig.~\ref{fig:shadowconcreteentity}, because -- according to ChatGPT -- it is an observable and tangible aspect of the physical world. However, in Fig.~\ref{fig:shadownotconcreteentity} exactly the same question is answered oppositely, because -- according to ChatGPT -- shadows while observable, they are not objects.  One possible explanation for the different answers is that in Fig.~\ref{fig:shadowconcreteentity}  `entity' is used by ChatGPT for the most generic category in the ontology, but in Fig.~\ref{fig:shadownotconcreteentity}  it seems used as a synonym for `object'. However, explanations like these are a kind of semantic pareidolia: we cannot help to read texts by generative AI as if they had a communicative intent. However, the different answers to the same question are a purely stochastic phenomenon. 

 By choosing the settings of GPT 3.5 appropriately, one can ensure that the same prompts will always produce the same output. However, this does not really address the concern.  Because while it ensures that exactly the same prompt leads to the same output, even minute changes that do not change the meaning of the prompt may lead to inconsistent results. 
Logical inconsistent responses are not rare and not limited to the distinction between concrete and abstract entities. E.g., `litter' is sometimes classified by ChatGPT as an artifact\footnote{
\url{https://chat.openai.com/share/7477e9f5-b167-4c51-be2f-4e11386a5104}
 } and sometimes not\footnote{
\url{https://chat.openai.com/share/2c449233-04c0-4ffa-8e26-d2d76c9dda58}
 }. Interestingly, the inconsistent responses seem to be sometimes inherited along the taxonomies. E.g., in ChatGPT's TLO shadows are a kind of phenomenon. Phenomena are also sometimes classified as concrete and sometimes as abstract.

While in most cases the ChatGPT's responses are  consistent and do not change significantly if the prompts are rephrased,  
these examples illustrate that there is a significant number of cases where this is not the case. 
Thus, if one attempts to use LLMs for the purpose of ontology development, it is important to cross-validate the classifications of the LLMs with the help of a variety of prompts to ensure that the results are robust. Further, there is likely a significant number of cases where the LLM will provide inconsistent results, and, thus, one needs to account for that possibility in the design of the workflow.

\section{Conclusion}

There is a significant research interest in the potential use of LLMs in ontology engineering. The development of such tools would be easier, if the texts that are generated by LLMs share a top-level ontology, i.e., they are committed to the existence of the same kind of entities and share the same ontological assumptions. Thus, the first question is whether there is such a top-level ontology, and, if there is, what it looks like and whether it is suitable for the purpose of ontology engineers. In this paper, we investigated these questions as a case study for ChatGPT~3.5. 

As we discussed in Section~\ref{sec:methodology},  ChatGPT does not use a top-level ontology (TLO) in the sense of an OWL file that may be downloaded. However, there is a stable core of ontological commitments and assumptions, which are repeatable across many interactions with the system. In this sense there ChatGPT uses a TLO. In  Section~\ref{sec:topLevel} we presented the taxonomy that ontology and some important ontological assumptions.
(The taxonomy is also available as an OWL file.) 

ChatGPT's ontological hierarchy of categories contains some distinctions, which are familiar from the ontological literature. 
However, it differs significantly from popular existing top-level ontologies. This may be an obstacle to the use of LLM-based tools within ontology engineering projects, which reuse existing TLO (e.g., DOLCE or BFO). For example,  any effort to use the GPT 3.5 model to extend a BFO-based ontology will run into the issue, that GPT 3.5 uses a different ontological categorization than BFO. (The differences are so significant that no simple ontology alignment is possible.) 
Thus, without some effort to ensure compatibility with BFO (e.g., by choosing the prompts accordingly), the responses of the LLM will be incompatible with the existing BFO-based taxonomy of the ontology, and, therefore, the result of the ontology extension is going to be incoherent and, possibly, even logically inconsistent. 

A different, and more challenging issue is the fact that LLMs do not learn to disambiguate terms, but rather to use terms appropriately in a given context. As we discussed, this results in a kind of `ontological overload', where entities a classified as members of different, supposedly disjoint categories. For ChatGPT the same entity may be abstract and concrete, depending on the context. 
Further, minute changes to the prompt or even the same prompt may lead to contradictory results. Thus, any attempt to use LLMs for ontology development needs to involve some strategy to manage `ontological overload' and inconsistent responses, otherwise, the resulting ontology will be of poor quality.  
Any good ontology provides clear and unambiguous definitions, which enable the consistent usage of its terms. On their own, LLMs deliver neither.

\bibliography{bibliography}

\end{document}